\documentclass[11pt]{article}
\usepackage{acl-hlt2011}
\usepackage{times}
\usepackage{latexsym}
\usepackage{amsmath}
\usepackage{multirow}
\usepackage{graphicx}
\usepackage{lscape}
\usepackage{url}
\usepackage{mdwlist}

\newcommand\T{\rule{0pt}{2.6ex}}
\newcommand\B{\rule[-1.2ex]{0pt}{0pt}}

\title{Perception of Personality and Naturalness through Dialogues by\\ 
Native Speakers of American English and Arabic}

\author{Maxim Makatchev\\
  Robotics Institute \\
  Carnegie Mellon University\\
  Pittsburgh, PA, USA \\
  {\tt mmakatch@cs.cmu.edu} \\\And
  Reid Simmons \\
  Robotics Institute \\
  Carnegie Mellon University\\
  Pittsburgh, PA, USA \\
  {\tt reids@cs.cmu.edu} \\}

\date{}

\begin{document}
\maketitle
\begin{abstract}
Linguistic markers of personality traits have been studied extensively, but few cross-cultural studies exist. In this paper, we evaluate how native speakers of American English and Arabic perceive personality traits and naturalness of English utterances that vary along the dimensions of verbosity, hedging, lexical and syntactic alignment, and formality. The utterances are the turns within dialogue fragments that are presented as text transcripts to the workers of Amazon's Mechanical Turk. The results of the study suggest that all four dimensions can be used as linguistic markers of all personality traits by both language communities. A further comparative analysis shows cross-cultural differences for some combinations of measures of personality traits and naturalness, the dimensions of linguistic variability and dialogue acts. 
\end{abstract}

\section{Introduction}
\label{sec:Intro}

English has been used as a lingua franca across the world, but the usage differs. The variabilities in English introduced by dialects, cultures, and non-native speakers result in different syntax and words expressing similar meanings and in different meanings attributed to similar expressions. These differences are a source of \textsl{pragmatic failures} \cite{Thomas1983}: situations when listeners perceive meanings and affective attitudes unintended by speakers. For example, \newcite{Thomas1984} reports that usage of Illocutionary Force Indicating Devices (IFIDs, such as ``I warn you'', \cite{searle1969}) in English by native speakers of Russian causes the speakers to sometimes appear ``inappropriately domineering in interactions with English-speaking equals.''
  Dialogue systems, just like humans, may misattribute attitudes and misinterpret intent of  user's utterances. Conversely, they may also cause misattributions and misinterpretations on the user's part. Hence, taking into account the user's dialect, culture, or native language may help reduce pragmatic failures.

This kind of adaptation requires a mapping from utterances, or more generally, their linguistic features, to meanings and affective attributions for each of the target language communities. In this paper we present an exploratory study that evaluates such a mapping from the linguistic features of verbosity, hedging, alignment, and formality (as defined in Section~\ref{Sec:Stimuli}) to the perceived personality traits and naturalness across the populations of native speakers of American English and Arabic. 

Estimating the relationship between linguistic features and their perception across language communities faces a number of methodological difficulties.
First, language communities shall be outlined, in a way that will afford generalizing within their populations. Defining language communities is a hard problem, even if it is based on the ``mother tongue''~\cite{Laitin2000}. 
Next, linguistic features that are potentially important for the adaptation must be selected. These are, for example, the linguistic devices that contribute to realization of \textsl{rich points}~\cite{AgarBook1994}, i.e. the behaviors that signal differences between language communities. To be useful for dialogue system research, the selected linguistic features should be feasible to implement in natural language generation and interpretation modules. Then, a corpus of stimuli that span the variability of the linguistic features must be created. The stimuli should reflect the context where the dialogue system is intended to be used. For example, in case of an information-giving dialogue system, the stimuli should include some question-answer adjacency pairs~\cite{Sacks1973}. Finally, scales should be chosen to allow for scoring of the stimuli with respect to the metrics of interest. These scales should be robust to be applied within each of the language communities.

In the remainder of this paper, we describe each of these steps in the context of an exploratory study that evaluates perception of English utterances by native speakers of American English and Arabic. Our application is an information-giving dialogue system that is used by the robot receptionists (roboceptionists) in Qatar and the United States~\cite{makatchev2009,hala_patterns2010}. In the next section, we continue with an overview of the related work. Section~\ref{Sec:Experiment} introduces the experiment, including the selection of stimuli, measures, design, and describes the recruitment of participants via Amazon's Mechanical Turk (MTurk). We discuss results in Section~\ref{Sec:Results} and provide a conclusion in Section~\ref{Sec:Conclusion}.



\section{Related work}
\label{Sec:RelatedWork}

\subsection{Cross-cultural variability in English}
Language is tightly connected with culture~\cite{AgarBook1994}. As a result, even native speakers of a language use it differently across dialects (e.g. African American Vernacular English and Standard American English), genders (see, for example, \cite{Lakoff1973}) and social statuses (e.g.~\cite{Huspek1989}), among other dimensions. 

Speakers of English as a second language display variabilities in language use that are consistent with their native languages and backgrounds. For example, \newcite{Nelson1996} reports that Syrian speakers of Arabic tend to use different compliment response strategies as compared with Americans. \newcite{Aguilar1998} reviews types of pragmatic failures that are influenced by native language and culture. In particular, he cites \newcite{Davies1987} on a pragmatic failure due to \textsl{non-equivalence of formulas}: native speakers of Moroccan Arabic use a spoken formulaic expression to wish a sick person quick recovery, whereas in English the formula ``get well soon'' is not generally used in speech. \newcite{Feghali1997} reviews features of Arabic communicative style, including indirectness (concealment of wants, needs or goals \cite{Gudykunst1988}), elaborateness (rich and expressive language use, e.g. involving rhetorical patterns of exaggeration and assertion~\cite{Patai1983}) and affectiveness (i.e. ``intuitive-affective style of emotional appeal''~\cite{Glenn1977}, related to the patterns of organization and presentation of arguments). 

In this paper, we are concerned with English usage by native speakers of American English and native speakers of Arabic. We have used the features of the Arabic communicative style  outlined above as a guide in selecting the dimensions of linguistic variability that are presented in Section~\ref{Sec:Stimuli}.


\subsection{Measuring pragmatic variation}

Perception of pragmatic variation of spoken language and text has been shown to vary across cultures along the dimensions of  personality (e.g.~\cite{Sherer1972}), emotion (e.g.~\cite{Burkhardt2006}), deception (e.g.~\cite{Bond1990}), among others. Within a culture, personality traits such as extraversion, have been shown to have consistent markers in language (see overview in \cite{MairesseWalkerMehl2007}). For example, \newcite{Furnham1990} notes that in conversation, extraverts are less formal and use more verbs, adverbs and pronouns.  However, the authors are not aware of any quantitative studies that compare linguistic markers of personality across cultures. The present study aims to help fill this gap.

A mapping between linguistic dimensions and personality has been evaluated by grading essays and conversation extracts~\cite{MairesseWalkerMehl2007}, and by grading utterances generated automatically with a random setting of linguistic parameters~\cite{Mairesse2008}. In the exploratory study presented in this paper, we ask our participants to grade dialogue fragments that were manually created to vary along each of the four linguistic dimensions (see Section~\ref{Sec:Stimuli}).

\section{Experiment}
\label{Sec:Experiment}

In the review of related work, we presented some evidence supporting the claim that linguistic markers of personality may differ across cultures. In this section, we describe a study that evaluates perception of personality traits and naturalness of utterances by native speakers of American English and Arabic.
\subsection{Stimuli}
\label{Sec:Stimuli}

The selection of stimuli attempts to satisfy three objectives. First, our application: our dialogue system is intended to be used on a robot receptionist. Hence, the stimuli are snippets of dialogue that include four dialogue acts that are typical in this kind of embodied information-giving dialogue \cite{makatchev2009}: a greeting, a question-answer pair, a disagreement (with the user's guess of an answer), and an apology (for the robot not knowing the answer to the question).

Second, we would like to vary our stimuli along the linguistic dimensions that are potentially strong indicators of personality traits. Extraverts, for example, are reported to be more verbose (use more words per utterances and more dialogue turns to achieve the same communicative goal), less formal \cite{Furnham1990} (in choice of address terms, for example), and less likely to hedge (use expressions such as ``perhaps'' and ``maybe'')~\cite{Nass1995}. Lexical and syntactic alignment, namely, the tendency of a speaker to use the same lexical and syntactic choices as their interlocutor, is considered, at least in part, to reflect the speaker's co-operation and willingness to adopt the interlocutor's perspective~\cite{Haywood2003}. There is some evidence that the degree of alignment is associated with personality traits of the speakers~\cite{Gill2004}.

Third, we would like to select linguistic dimensions that potentially expose cross-cultural differences in perception of personality and naturalness. In particular, we are interested in the linguistic devices that help realize \textsl{rich points} (the behaviors that signal differences) between the native speakers of American English and Arabic. We choose to realize indirectness and elaborateness, characteristic of Arabic spoken language~\cite{Feghali1997}, by varying the dimensions of verbosity and hedging. High \textsl{power distance}, or influence of relative social status on the language~\cite{Feghali1997}, can be realized by the degrees of formality and alignment.

In summary, the stimuli are dialogue fragments where utterances of one of the interlocutors vary across (1) dialogue acts: a greeting, question-answer pair, disagreement, apology, and (2) four linguistic dimensions: verbosity, hedging, alignment, and formality. Each of the linguistic dimensions is parameterized by 3 values of valence: negative, neutral and positive. Within each of the four dialogue acts, stimuli corresponding to the neutral valences are represented by the same dialogue across all four linguistic dimensions.
The four linguistic dimensions are realized as follows: 
\begin{itemize*}
\item Verbosity is realized as number of words within each turn of the dialogue. In the case of the greeting, positive verbosity is realized by increased number of dialogue turns.\footnote{The multi-stage greeting dialogue was developed via ethnographic studies conducted at Alelo by Dr. Suzanne Wertheim. Used with permission from Alelo, Inc.}
\item Positive valence of hedging implies more tentative words (``maybe,'' ``perhaps,'' etc.) or expressions of uncertainty (``I think,'' ``if I am not mistaken''). Conversely, negative valence of hedging is realized via words ``sure,'' ``definitely,'' etc.
\item Positive valence of alignment corresponds to preference towards the lexical and syntactic choices of the interlocutor. Conversely, negative alignment implies less overlap in lexical and syntactic choices between the interlocutors.
\item Our model of formality deploys the following linguistic devices: in-group identity markers that target positive face~\cite{BrownLevinson} such as address forms, jargon and slang, and deference markers that target negative face, such as ``kindly'', terms of address, hedges. These devices are used in Arabic politeness phenomena~\cite{Farahat2009}, and there is an evidence of their pragmatic transfer from Arabic to English (e.g.~\cite{Bardovi2007} and \cite{Ghawi1993}). 
\end{itemize*}
The complete set of stimuli is shown in Tables~\ref{Table:InventoryControl}--\ref{Table:InventoryFormality}. 

Each dialogue fragment is presented as a text on an individual web page. On each page, the participant is asked to imagine that he or she is one of the interlocutors and the other interlocutor is described as ``a female receptionist in her early 20s and of the same ethnic background'' as that of the participant. The description of the occupation, age, gender and ethnicity of the interlocutor whose utterances the participant is asked to evaluate should provide minimal context and help avoid variability due to the implicit assumptions that subjects may make.

\subsection{Measures}
\label{Sec:Measures}

In order to avoid a possible interference of scales, we ran two versions of the study in parallel. In one version, participants were asked to evaluate the receptionist's utterances with respect to measures of the Big Five personality traits~\cite{John1999}, namely the traits of extraversion, agreeableness, conscientiousness, emotional stability, and openness, using the ten-item personality questionnaire (TIPI, see \cite{Gosling2003}). In the other version, participants were asked to evaluate the receptionist's utterances with respect to their naturalness on a 7-point Likert scale by answering the question ``Do you agree that the receptionist's utterances were natural?'' The variants of such a naturalness scale were used by \newcite{Burkhardt2006} and \newcite{Mairesse2008}.

\subsection{Experimental design}
\label{Sec:Design}

The experiment used a crossed design with the following factors: dimensions of linguistic variability  (verbosity, hedging, alignment, or formality), valence (negative, neutral, or positive), dialogue acts (greeting, question-answer, disagreement, or apology), native language (American English or Arabic) and gender (male or female).

In an attempt to balance the workload of the participants, depending on whether the participant was assigned to the study that used personality or naturalness scales, the experimental sessions consisted of one or two linguistic variability conditions---12 or 24 dialogues respectively.  Hence valence and dialogue act were within-subject factors, while linguistic variability dimension were treated as an across-subject factor, as well as native language and gender. Within each session the items were presented in a random order to minimize possible carryover effects.

\subsection{Participants}
\label{Sec:Subjects}

We used Amazon's Mechanical Turk (MTurk) to recruit native speakers of American English from the United States and native speakers of Arabic from any of the set of predominantly Arabic-speaking countries (according to the IP address). 

Upon completion of each task, participants receive monetary reward as a credit to their MTurk account. Special measures were taken to prevent multiple participation of one person in the same  study condition: the study website access would be refused for such a user based on the IP address, and MTurk logs were checked for repeated MTurk user names to detect logging into the same MTurk account from different IP addresses. Hidden questions were planted within the study to verify the fluency in the participant's reported native language.

The distribution of the participants across countries is shown in Table~\ref{table:dem_country}.
We observed a regional gender bias similar to the one reported by \newcite{Ross2010}: there were 100 male and 55 female participants in the Arabic condition, and 63 male and 103 female participants in the American English condition.

\begin{table}[bt]
\centering
{\small
\begin{tabular}{|l|l|c|}
\hline
Language \T \B & Country & $N$\\
\hline\hline
Arabic \T & Algeria & 1 \\
& Bahrain & 1\\
& Egypt & 56 \\
& Jordan & 32 \\
& Morocco & 45 \\
& Palestinian Territory & 1\\
& Qatar & 1\\
& Saudi Arabia & 5 \\
& United Arab Emirates & 13 \\
\hline
& Total & 155\\
\hline
American English & United States & 166 \\
\hline
\end{tabular}
}
\caption[Study participants by country]{Distribution of study participants by country.}\label{table:dem_country}
\end{table}

\section{Results}
\label{Sec:Results}

We analyzed the data by fitting linear mixed-effects (LME) models~\cite{PinheiroBates2000} and performing  model selection using ANOVA. The comparison of models fitted to explain the personality and naturalness scores (controlling for language and gender), shows significant main effects of valence and dialogue acts for all pairs of personality traits (and naturalness) and linguistic features. The results also show that for every personality trait (and naturalness) there is a linguistic feature that results in a significant three-way interaction between its valence, the native language, and the dialogue act. These results suggest that (a) for both language communities, every linguistic dimension is associated with every personality trait and naturalness, for at least some of the dialogue acts, (b) there are differences in the perception of every personality trait and naturalness between the two language communities.


To further explore the latter finding, we conducted a post-hoc analysis consisting of paired t-tests that were performed pairwise between the three values of valence for each combination of language, linguistic feature, and personality trait (and naturalness).
Note, that comparing raw scores between the language conditions would be prone to find spurious differences due to potential culture-specific tendencies in scoring on the Likert scale: (a) perception of magnitudes and (b) appropriateness of the intensity of agreeing or disagreeing. Instead, we compare the language conditions with respect to (a) the relative order of the three valences and (b) the binarized scores, namely whether the score is above 4 or below 4 (with scores that are not significantly different from 4 excluded from comparison), where 4 is the neutral point of the 7-point Likert scale. 
 
\begin{figure*}[bthp]
         \includegraphics[width=\linewidth]{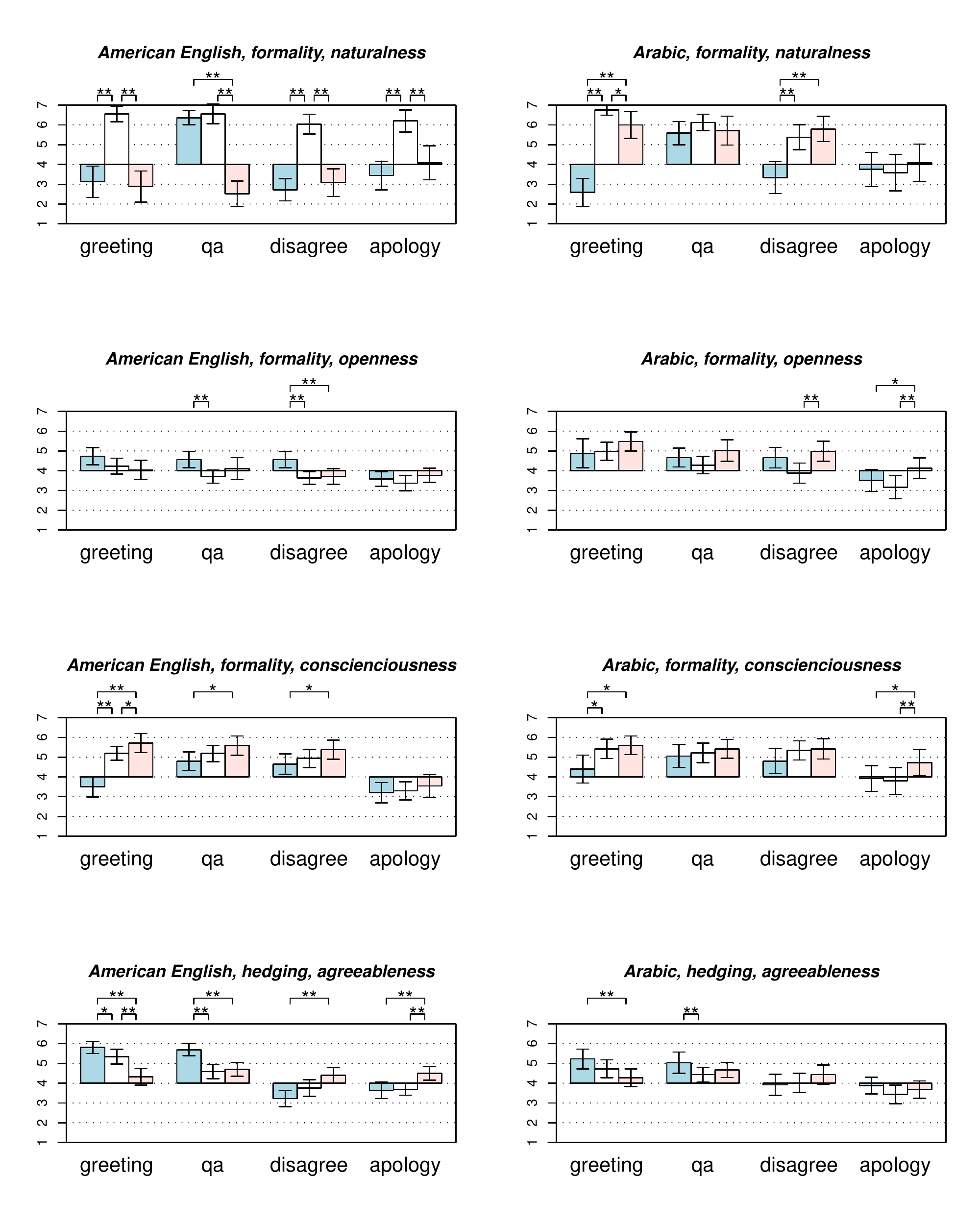}
        \caption[Selected stuff]{A subset of data comparing scores on the Big Five personality traits and naturalness as given by native speakers of American English (left half of the page) and Arabic (right half of the page). Blue, white, and pink bars correspond to negative, neutral, and positive valences of the linguistic features respectively. Dialogue acts listed along the horizontal axis are a greeting, question-answer pair, disagreement, and apology. Error bars the 95\% confidence intervals, brackets above the plots correspond to p-values of paired t-tests at significance levels of 0.05 (\textasteriskcentered) and 0.01 (\textasteriskcentered\textasteriskcentered)  after Bonferroni correction.}\label{fig:selected_abs}
\end{figure*}

The selected results of the post-hoc analysis are shown in Figure~\ref{fig:selected_abs}. The most prominent cross-cultural differences were found in the scoring of naturalness across the valences of the formality dimension. Speakers of American English, unlike the speakers of Arabic, find formal utterances unnatural in greetings, question-answer and disagreement dialogue acts. Formal utterances tend to also be perceived as indicators of openness and conscientiousness by Arabic speakers, and not by American English speakers, in disagreements and apologies respectively. Finally, hedging in apologies is perceived as an indicator of agreeableness by American English speakers, but not by speakers of Arabic.

Interestingly, no qualitative differences across language conditions were found in the perception of extraversion and stability. It is possible that this cross-cultural consistency confirms the view of the extraversion, in particular, as one of most consistently identified dimensions (see, for example, \cite{Gill2002}). It could also be possible that our stimuli were unable to pinpoint the extraversion-related rich points due to a choice of the linguistic dimensions or particular wording chosen. A larger variety of stimuli per condition, and an ethnography to identify potentially culture-specific linguistic devices of extraversion, could shed the light on this issue. 

\section{Conclusion}
\label{Sec:Conclusion}

We presented an exploratory study to evaluate a set of linguistic markers of Big Five personality traits and naturalness across two language communities: native speakers of American English living in the US, and native speakers of Arabic living in one of the predominantly Arabic-speaking countries of North Africa and Middle East. The results suggest that the four dimensions of linguistic variability are recognized as markers of all five personality traits by both language communities. A comparison across language communities uncovered some qualitative differences in the perception of openness, conscientiousness, agreeableness, and naturalness.

The results of the study can be used to adapt natural language generation and interpretation to native speakers of American English or Arabic. 
This exploratory study also supports the feasibility of the crowdsourcing approach to validate the linguistic devices that realize rich points---behaviors that signal differences across languages and cultures. 

Future work shall evaluate effects of regional dialects and address the issue of particular wording choices by using multiple stimuli per condition.



\section*{Acknowledgments}

This publication was made possible by the support of an NPRP grant from the Qatar National Research Fund. The statements made herein are solely the responsibility of the authors. 

The authors are grateful to Ameer Ayman Abdulsalam, Michael Agar, Hatem Alismail, Justine Cassell, Majd Sakr, Nik Melchior, and Candace Sidner for their comments on the study. 

\bibliographystyle{acl} 
\bibliography{nl}

\begin{landscape}
\begin{table}[bth]
\begin{tabular}{|p{2.5cm}|p{4.6cm}|p{4.3cm}|p{4.3cm}|p{4.3cm}|}
\hline
\T Dimension and valence \B & Greeting & Question-Answer & Disagreement & Apology\\
\hline\hline
\T Control\newline 
(neutral) & A: Good morning. \newline 
B: Good morning. How may I help you? &  
A: Could you tell me where the library is? \newline 
B: It's at the end of the hallway on your left. & 
A: Could you tell me where the library is? \newline
B: It's on the second floor. \newline
A: I thought it was on the first floor.\newline
B: No, there is no library on the first floor. \B & 
A: Could you tell me where the library is? \newline
B: Sorry, I don't know.\\
\hline
\end{tabular}
\caption[Stimuli for the control condition]{Stimuli corresponding to the neutral control condition, shared by all dimensions of linguistic variability.}\label{Table:InventoryControl}
\end{table}
\end{landscape}

\begin{landscape}
\begin{table}[bth]
\begin{tabular}{|p{2.5cm}|p{4.6cm}|p{4.3cm}|p{4.3cm}|p{4.3cm}|}
\hline
\T Dimension and valence \B & Greeting & Question-Answer & Disagreement & Apology\\
\hline\hline
\T Verbosity \newline
negative & 
A: Good morning. \newline
B: Morning. May I help you? & 
A: Could you tell me where the library is? \newline
B: End of the hallway on your left. &
A: Could you tell me where the library is? \newline
B: Second floor. \newline
A: I thought it was on the first floor. \newline
B: No, it is not. \B &
A: Could you tell me where the library is? \newline
B: I don't know. \\
\hline
\T Verbosity \newline
positive &
A: Good morning. \newline
B: Good morning. How are you today? \newline
A: I am doing well, thanks. You? \newline
B: Very well, thank you. How's your family? \newline
A: Everyone is doing fine, thanks. How about yours? \newline
B: Mine is doing well too. How may I help you? \B &
A: Could you tell me where the library is? \newline
B: Yes, just follow this hallway until it ends and you will find the library on 
your left hand side. &
A: Could you tell me where the library is? \newline
B: Yes, the library is on the second floor. \newline
A: I thought it was on the first floor. \newline
B: No, there is no library on the first floor of this building. &
A: Could you tell me where the library is? \newline
B: I am so sorry, I don't really know where the library is in this building.\\
\hline
\end{tabular}
\caption[Stimuli for the verbosity section]{Stimuli along the verbosity dimension.}\label{Table:InventoryVerbosity}
\end{table}
\end{landscape}

\begin{landscape}
\begin{table}[bth]
\begin{tabular}{|p{2.5cm}|p{4.3cm}|p{4.3cm}|p{4.6cm}|p{4.3cm}|}
\hline
\T Dimension and valence \B & Greeting & Question-Answer & Disagreement & Apology\\
\hline\hline
\T Hedging \newline
negative & 
 A: How are you? \newline
B: Definitely good, and you? \newline
A: Good, thank you. \newline 
B: How may I help you? &
A: Could you tell me where the library is? \newline
B: Sure, it is at the end of the hallway on your left. &
A: Could you tell me where the library is? \newline
B: The library is on the second floor. \newline
A: I thought it was on the first floor. \newline
B: No, there is definitely no library on the first floor. \B &
A: Could you tell me where the library is? \newline
B: Sorry, I have no idea where it is. \\
\hline
\T Hedging \newline
positive &
A: How are you? \newline
B: Good, I guess, and you? \newline
A: Good, thank you. \newline
B: Could I help you? &
A: Could you tell me where the library is? \newline
B: I think it is at the end of the hallway on your left.&
A: Could you tell me where the library is? \newline
B: I think the library is on the second floor. \newline
A: I thought it was on the first floor. \newline
B: No, if I am not mistaken, there is no library on the first floor. \B &
A: Could you tell me where the library is? \newline
B: Sorry, I am not sure. \\
\hline
\end{tabular}
\caption[Stimuli for hedging section of the experiment]{Stimuli along the hedging dimension.}\label{Table:InventoryHedging}
\end{table}
\end{landscape}

\begin{landscape}
\begin{table}[bth]
\begin{tabular}{|p{2.5cm}|p{4.3cm}|p{4.3cm}|p{4.6cm}|p{4.3cm}|}
\hline
\T Dimension and valence \B & Greeting & Question-Answer & Disagreement & Apology\\
\hline\hline
\T Alignment \newline
negative &
A: How are you? \newline
B: Good morning. How may I help you? &
A: Could you tell me where the bathroom is? \newline
B: The restroom is at the end of the hallway on your left. &
A: Could you tell me where the bathroom is? \newline
B: The restroom is on the second floor. \newline
A: I thought the bathroom was on the first floor. \newline
B: No, there is no restroom on the first floor. \B &
A: Could you tell me where the bathroom is? \newline
B: Sorry, I don't know about the restroom. \\
\hline
\T Alignment \newline
positive &
A: How are you? \newline
B: Good, how are you? How may I help you? &
A: Could you tell me where the bathroom is? \newline
B: The bathroom is at the end of the hallway on your left. &
A: Could you tell me where the bathroom is? \newline
B: The bathroom is on the second floor. \newline
A: I thought the bathroom was on the first floor. \newline
B: No, there is no bathroom on the first floor. \B &
A: Could you tell me where the bathroom is? \newline
B: Sorry, I don't know where the bathroom is. \\
\hline
\end{tabular}
\caption[Stimuli for lexical and syntactic alignment section of the experiment]{Stimuli along the lexical and syntactic alignment dimension.}\label{Table:InventoryAlignment}
\end{table}
\end{landscape}

\begin{landscape}
\begin{table}[bth]
\begin{tabular}{|p{2.5cm}|p{4.3cm}|p{4.3cm}|p{4.6cm}|p{4.3cm}|}
\hline
\T Dimension and valence \B & Greeting & Question-Answer & Disagreement & Apology\\
\hline\hline
\T Formality \newline
negative &
A: Good morning.\newline
B: What's up? Need anything? &
A: Could you tell me where the library is? \newline
B: Just go to the end of the hallway, you can't miss it. &
A: Could you tell me where the library is? \newline
B: Go to the second floor. \newline
A: I thought it was on the first floor. \newline
B: No, honey, there is none on the first floor. \B &
A: Could you tell me where the library is? \newline
B: Sorry about that, I have no idea. \\
\hline
\T Formality \newline
positive &
A: Good morning. \newline
B: Good morning, sir (madam). Would you allow me to help you with anything? &
A: Could you tell me where the library is? \newline
B: Kindly follow this hallway and you will encounter the entrance on your left. &
A: Could you tell me where the library is? \newline
B: Yes, you may find the library on the second floor. \newline
A: I thought it was on the first floor. \newline
B: I am afraid that is not correct, there is no library on the first floor. \B &
A: Could you tell me where the library is? \newline
B: I have to apologize, but I don't know.\\
\hline
\end{tabular}
\caption[Stimuli for formality section of the experiment]{Stimuli along the formality dimension.}\label{Table:InventoryFormality}
\end{table}
\end{landscape}

\end{document}